%% file: main.tex
\documentclass[letterpaper, 10 pt, conference]{ieeeconf} 
\IEEEoverridecommandlockouts
\usepackage{graphics} 
\usepackage{epsfig} 
\usepackage{amsmath} 
\usepackage{amssymb} 
\usepackage{color}
\usepackage{bm}
\usepackage{subcaption}
\usepackage{multirow}
\usepackage{array}
\usepackage{booktabs}
\usepackage{algorithm,algcompatible}
\usepackage{mathtools}
\usepackage{cite}
\usepackage{bbding}
\usepackage{cuted}
\definecolor{pointcolor}{RGB}{0,114,189}
\definecolor{linecolor}{RGB}{217,83,25}
\definecolor{planecolor}{RGB}{237,177,32}
\definecolor{spherecolor}{RGB}{119,172,48}
\definecolor{ellipsoidcolor}{RGB}{126,47,142}
\definecolor{cylindercolor}{RGB}{77,190,238}
\definecolor{conecolor}{RGB}{162,20,47}

\graphicspath{{figures/}}

\title{\huge \bf Attention-Enhanced Cross-modal Localization Between 360 Images and Point Clouds}

\author{Zhipeng Zhao\quad  Huai Yu\quad  Chenwei Lyu\quad  Wen Yang \quad Sebastian Scherer 
\thanks{Zhipeng Zhao, Huai Yu, Chenwei Lyu, Wen Yang are with the Electronic Information School, Wuhan University, Wuhan, Hubei 430072, China. {\tt\small\{zhaozhp,yuhuai,lvchenwei,yangwen\}@whu.edu.cn}}
\thanks{Sebastian Scherer is with Robotics Institute, Carnegie Mellon University, Pittsburgh, PA 15213, USA. {\tt\small basti@andrew.cmu.edu}}
}

\begin{document}

\maketitle
\input{sections/abstract}
\input{sections/intro}
\input{sections/related}
\input{sections/methods}
\input{sections/experiments}

\input{sections/conclusion}

\bibliographystyle{IEEEtran}
\bibliography{references}

\end{document}

%% file: sections/abstract.tex
\begin{abstract}
Visual localization plays an important role for intelligent robots and autonomous driving, especially when the accuracy of GNSS is unreliable. Recently, camera localization in LiDAR maps has attracted more and more attention for its low cost and potential robustness to illumination and weather changes. However, the common used pinhole camera has a narrow \emph{Field-of-View}, thus leading to limited information compared with the omni-directional LiDAR data. To overcome this limitation, we focus on correlating the information of 360 equirectangular images to point clouds, proposing an end-to-end learnable network to conduct cross-modal visual localization by establishing similarity in high-dimensional feature space. Inspired by the attention mechanism, we optimize the network to capture the salient feature for comparing images and point clouds. We construct several sequences containing 360 equirectangular images and corresponding point clouds based on the KITTI-360 dataset and conduct extensive experiments. The results demonstrate the effectiveness of our approach.
\end{abstract}

%% file: sections/intro.tex
\section{Introduction}
Locating the position of \emph{an image} in \emph{the point cloud map} is of great importance for mobile robots and autonomous vehicles with numerous applications such as Simultaneous Localization and Mapping (SLAM) \cite{newman2005slam} and Virtual Reality \cite{yang2020mobile3drecon}. Previous work of place recognition based on image-to-image retrieval is sensitive to illumination and weather changes \cite{tang2015neighborhood,arandjelovic2016netvlad}, while laserscan-pointclouds matching \cite{uy2018pointnetvlad} is fragile to geometric degeneracy and wide open area. Image to 3D map matching using lightweight and cheap cameras is robust to illumination and weather changes. Moreover, the 360 camera has an omni-directional view and its panoramic information exactly corresponds to the omni-directional LiDAR point clouds, which makes it more suitable for localization in point cloud maps. However, this task remains challenging due to the cross-modal heterogeneity gap and the serious distortion of 360 images.

\begin{figure}[t]
    \centering
    \includegraphics[width=\linewidth]{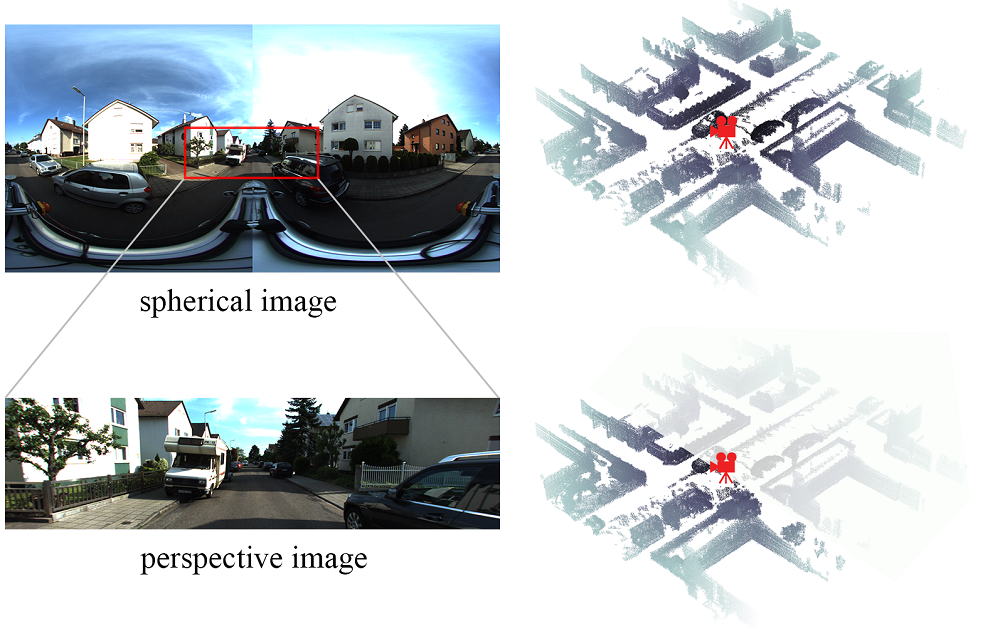}
    \caption{Comparison of spherical images and perspective images with point cloud counterparts. The \emph{Right} side shows the point clouds corresponding to the images on the \emph{Left}, which were obtained at the same location.} \vspace{-3mm}
    \label{fig:intro}
\end{figure}

The information from 2D images and 3D point clouds is heterogeneous. To illustrate, images capture the objects' texture information, while point clouds record a set of points in the space reflecting geometric characteristics. Therefore, the features generated by different modal data are different in description and located in disparate subspaces \cite{guo2019deep}, which significantly increases the difficulty of similarity measure compared with image-to-image or laserscan-to-pointcloud retrieval. The common approach to bridge the heterogeneity gap is to project multi-modal feature descriptors into a shared place. Researchers have recently proposed several methods to match images and point clouds based on their descriptors such as keypoints \cite{feng20192d3d} and aggregated local features \cite{cattaneo2020global}. Cattaneo \emph{et al.} \cite{cattaneo2020global} map different modal data into a shared embedding space by two DNNs, one for images and another for point clouds, which extract features of the data based on CNN and then aggregates all the features into descriptors. However, the perspective images used in current methods suffer from limited field-of-view, which can only depict information from a limited angle in a fixed direction. Besides, many LiDAR points fail to be utilized when establishing correspondence between images and point clouds, as shown in Fig.\ref{fig:intro}. To obtain more comprehensive visual information, 360 cameras have attracted increasing attention in robotic systems. But, with its natural sphere surface, the projected 360 image inevitably has significant distortions \cite{courbon2007generic} \cite{Coors2018ECCV}. For instance, the same objects at different latitudes of the sphere surface differ greatly from different viewpoints, thus it makes the feature extraction used in current methods less effective. Additionally, some structures are more informative for the recognition, however, current methods often give equal weight to all local features. Therefore, how to match 360 images and point clouds in cross-modal localization is still an open problem.


\begin{figure}[t]
    \centering
    \includegraphics[width=\linewidth]{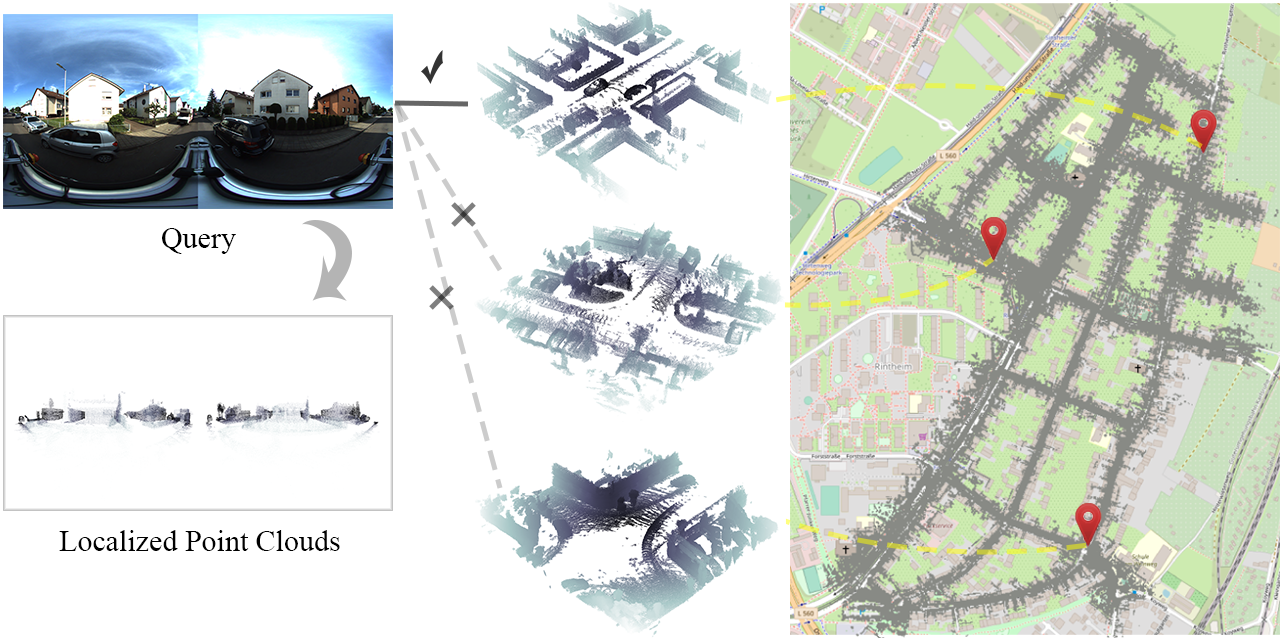}
    \caption{A schematic of the cross-modal localization. The localization is performed by comparing the query 360 image with the point clouds sub-maps from the global map and then finding the closest sub-map to determine the location.} \vspace{-3mm}
    \label{fig:intro_method}
\end{figure}

In this paper, we propose a cross-modal retrieval model between 360 images and point clouds which targets the camera localization in LiDAR maps. The framework can leverage the spherical image and establish connections between information from different modalities by learning representations for each modality. Firstly, with richer 360 information by dual-fisheye image, mobile robots can better perceive the surrounding environment since omni-directional images can provide more information about the scene, such as objects from the back of robots. Besides, spherical images are more robust to orientation changes and can fully utilize the omni-directional LiDAR data. Secondly, the model mitigates the influence of 360 image distortion by spherical-oriented convolution improvement. Furthermore, to better establish the correspondence between high-dimensional features, we intend to enable the model to focus on salient features of each modality that can be used for comparison between the different modalities.

Our main contribution can be summarized as:
\begin{itemize}
    \item We propose an approach of cross-modal localization between 360 spherical images and point clouds from a large-scale database, which relates panoramic information between these two modalities. 
    \item To address the heterogeneity gap between images and point clouds, we propose to apply the attention mechanism in the cross-modal retrieval model to find out the specific features used to link the two modalities, which improves the effectiveness of cross-modal retrieval. 
    \item We construct a 2D-3D localization dataset consisting of equirectangular images and 3D LiDAR sub-maps based on the KITTI360 dataset \cite{liao2022kitti} with augmentation under different lighting and weather conditions or geometric transformations.
\end{itemize}

The rest of this paper is structured as follows: Section \ref{sec:related_work} discusses the prior work on visual localization. Section \ref{sec:approach} introduces our proposed approach including the network architecture and implementation details. In Section \ref{sec:experiments}, experiments on the public dataset are presented. Finally, conclusions are drawn in Section \ref{sec:conclusion}.

%% file: sections/related.tex
\section{Related Work}
\label{sec:related_work}
To tackle the problem of localization for the mobile robot, the common practice is to retrieve the best matching scene from a database containing global scene and location information using the currently acquired surrounding environment information (e.g. images or point clouds). It allows the robot to know its present position for pose estimation with the Perspective-n-Point \cite{haralick1991analysis} and geometric verification \cite{lee2014unsupervised}. According to the modality of the information employed in the retrieval, This section will review image-based, point cloud-based, and cross-modal-based work. 

\subsection{Image based retrieval}
\label{sec:image}
The key to the image-based task is to find out how to describe the image, thus, the common approaches consist of two steps: (1) feature extraction, followed by (2) feature aggregation. Traditionally, local features of the image are extracted by methods, such as SIFT \cite{lowe2004distinctive}, SURF \cite{Bay2006SURF} and ORB \cite{Rublee2011ORB}. Recently, convolution neural networks (CNNs), such as VGG \cite{simonyan2014very} and ResNet \cite{he2016deep}, are replacing handcrafted feature extraction and performing well on images. Additionally, for 360 images,  Cohen \emph{et al.} \cite{s.2018spherical} replace filter translations by rotations for cross-correlation and introduce the rotation-equivariant spherical CNNs. Coors \emph{et al.} \cite{Coors2018ECCV} focus on the representation learning for omni-directional images and reverse the distortions by wrapping the filters around the sphere. Likewise, previously used aggregation methods, such as bag of words \cite{nister2006scalable}, are currently replaced by deep learning techniques. Based on the Vector of Locally Aggregated Descriptors (VLAD) \cite{jegou2010aggregating}, Arandjelovic \emph{et al.} \cite{arandjelovic2016netvlad} propose a new generalized VLAD layer which is differentiable and can be plugged into any CNN architecture.

However, it is difficult to extract invariant features as images at the same location vary greatly under different lighting and seasonal conditions. For the DNN-based method, a large amount of image data under different conditions is required for training, especially for more complex scenes. In the practical application of robot positioning, it becomes complicated to make an effective database with images that can accommodate queries at any moment.

\subsection{Point cloud based retrieval}
Compared with images, point clouds are more robust in describing scenes under different lighting and seasonal conditions. Point clouds that preserve the structural features of scenes are increasingly used in mobile robots and autonomous driving today, and research related to the extraction and understanding of point cloud features has achieved good results. In \cite{qi2017pointnet}, Qi \emph{et al.} first propose a novel neural network, named PointNet, which can directly consume unordered points and extract permutation-invariant features for classification and segmentation tasks. On this basis, Uy \emph{et al.} \cite{uy2018pointnetvlad} propose an end-to-end trainable model, named PointNetVLAD, which combines the PointNet and NetVLAD to tackle point cloud based retrieval for place recognition. The reference map is divided into a series of sub-maps with approximately the same area of coverage to serve as a database for retrieval. Given a query point cloud, the task is to retrieve the most similar sub-maps from the database. To facilitate faster search computations, \cite{uy2018pointnetvlad} uses a fully connected layer after the VLAD to compress the VLAD descriptor into a compact feature vector. The experimental results reveal that the use of point clouds make the result more robust to 
spectral changes, compared with images, since point cloud based PointNetVLAD outperforms image based NetVLAD in the day-to-night retrieval.

However, different from the query submap in \cite{uy2018pointnetvlad}, which is cut from the global reference map created in advance, the scans acquired in real time using LiDAR during the actual robot movement contain relatively limited information describing the environment, which makes it difficult to perform point cloud based retrieval efficiently using onboard LiDAR scan information. In addition, for scenes with scarce geometric structures, such as tunnels and corridors, point clouds suffer from geometric degeneracy, reflecting their limitations in describing the wide open area. In this case, point clouds cannot effectively represent the scene. Furthermore, the price and weight of the LiDAR make it unsuitable for large-scale applications, especially for some small robots and lower-cost applications. Therefore, for many applications where positioning is required, it is more appropriate to use the camera as an onboard sensor to acquire images in real-time and retrieve them from a pre-built point cloud map. 

\subsection{Cross-modal based retrieval}
Similar to the retrieval process in the same modal, cross-modal retrieval also compares the representation of the query image with the representation of the point cloud submap in the database to find the best match. Therefore, the challenge of cross-modal localization lies in describing and correlating the information of two different modalities (image and point cloud in our work) in the retrieval. Based on the Triplet architecture, Feng \emph{et al.} \cite{feng20192d3d} propose a deep network, named 2D3D-MatchNet, to jointly learn the keypoint descriptors of the 2D and 3D keypoints extracted from the image and point cloud. Different from the previous detector of 2D SIFT \cite{lowe2004distinctive} and 3D ISS \cite{zhong2009intrinsic}, Cattaneo \emph{et al.} \cite{cattaneo2020global} propose the shared 2D-3D embedding space and exploit two DNNs (e.g. NetVLAD and PointNetVLAD) which are jointly trained to produce similar embedding vectors for the image and point cloud. The two DNNs are both composed of two parts: a DNN-based feature extractor, and an aggregation layer. For each DNN, although different feature extraction(e.g. EdgeConv \cite{wang2019dynamic} and SECOND \cite{yan2018second} for point clouds) and aggregation layers are tested in \cite{cattaneo2020global}, all local features extracted from images or point clouds are directly aggregated separately, which increases the difficulty of model learning and affects the final results. By first projecting points back to the keyframe, Yin \emph{et al.} \cite{yin2021i3dloc} introduce the approach of extracting cross-domain symmetric place descriptors to match equirectangular images to the 3D range projections. To eliminate condition-related factors, a Generative adversarial Network(GAN) is designed to extract condition-invariant features and a spherical convolution network is utilized to learn viewpoint-invariant symmetric descriptors. Besides, Lai \emph{et al.} \cite{lai2022adafusion} propose an adaptive weighting visual-LiDAR fusion method combining image and point clouds for localization. The model exploits a weight generation branch to learn the weights for both modalities, considering their contribution in different situations.

The aforementioned cross-modal work mainly deals with perspective images but rarely with 360 images that contain panoramic information except for \cite{yin2021i3dloc}. Motivated by the fact that fisheye images contain panoramic information more similar to point clouds, we focus on the cross-modal retrieval between 360 images and point clouds.


%% file: sections/methods.tex
\section{Approach}
\label{sec:approach}
\subsection{Overview}
Unlike localization by the same-modal retrieval, our work focuses on correlating image and point cloud information to obtain locations. Assuming that there is a global trajectory point cloud map $\mathcal{M}$ with known poses, it can be further divided into a series of sub-maps $m_i$, and these sub-maps with location information build up the \emph{database} $\mathcal{DB}=\left \{ m_1, m_2, \dots, m_N \right \}$. While the mobile robot or autonomous vehicle is in motion, the fisheye camera onboard captures 360 spherical images in real-time as a \emph{query}. Therefore, this problem can be defined as:

Given a 360 spherical image $\mathcal{I}$ as a \emph{query}, the goal is to find the sub-map $m_i$ at the same place from the \emph{database} to get the location where the image is taken.

To achieve this, information from spherical images and point clouds needs to be comparable with each other. More specifically, we aim to find a common representation space $V$ for images $\mathcal{I}$ and point clouds $m_i$, where images and point clouds from the same place are more similar than those from different places.

Thus metric learning of the model lies in learning two representation functions for two modalities:
\begin{equation}
    \mathbf{V}^I = f(\mathcal{I}), \mathbf{V}^P_i = g(m_i)
\end{equation}
where $\mathbf{V}^I \in \mathbb{R}^D $ is the representation of the image, $\mathbf{V}^P \in \mathbb{R}^D $ is the representation of the point cloud, and $D$ is the dimension of the representation space. The goal is to ensure that $\mathbf{V}^I$ and $\mathbf{V}^P_i$ are more similar than $\mathbf{V}^P_j$ when $m_i$ and $\mathcal{I}$ are in the same location, and $m_j$ is not.

\subsection{The Network Architecture}
\begin{figure}[t]
    \centering
    \includegraphics[width=\linewidth]{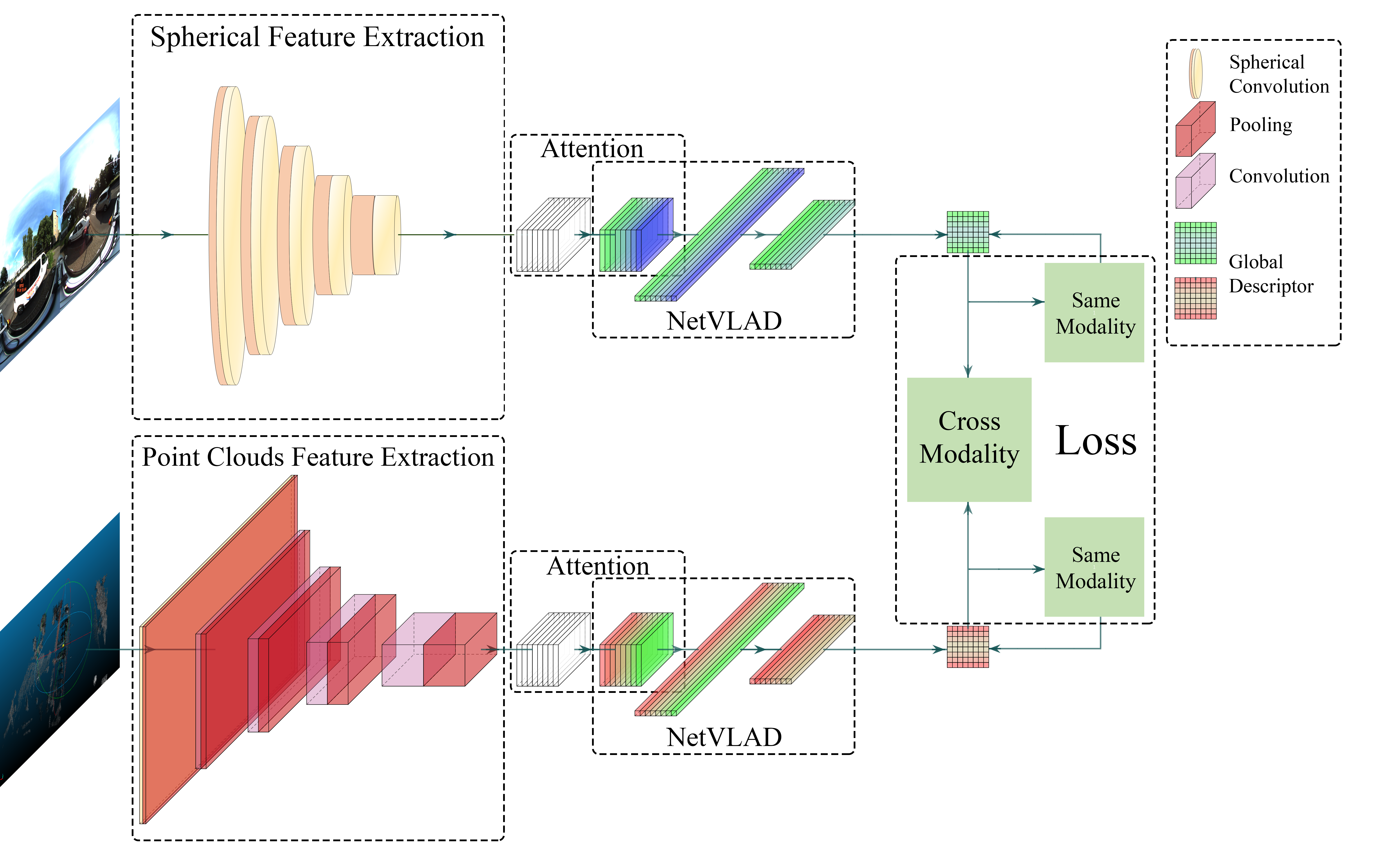}
    \caption{The Architecture of our Model for Cross-modal Localization. The inputs are the 360 image and the point cloud sub-map.} \vspace{-3mm}
    \label{fig:model}
\end{figure}

Although panoramic images and point cloud sub-maps at the same location contain similar information, it is difficult for the machine to identify the consistency of the scene directly from the original data due to the heterogeneity gap between them. Therefore, we propose a DNN-based network consisting of three main components: feature extraction, attention enhancement, and global description aggregation, as shown in Fig.\ref{fig:model}.

\subsubsection{Feature Extraction}
\paragraph{Image Feature Extraction} The input image is an \emph{equirectangular image} commonly used as a 360 image representation, where the latitude and longitude of the spherical image are mapped to horizontal and vertical grid coordinates \cite{Coors2018ECCV}. To solve the problem of distortion of the equirectangular image. We introduce the idea of spherical convolution, which encodes invariance to geometric transformations directly into the regular CNN \cite{s.2018spherical}. The method is to adjust the sampling grid locations of the filter when sampling the input image and intermediate feature maps. In this way, the filters are distorted like objects on the tangent plane of the sphere. The sampling area adjustment not only eliminates the influence of equirectangular image distortion on the sampling operation but also allows the filter to sample across the boundary of the 2D image, which enables the model to retain information about objects separated at the boundaries since the equirectangular image are continuous at the boundaries. In particular, based on ResNet-18 \cite{he2016deep} backbone, we mainly modified the sampling locations of the convolution and max pooling kernels to accommodate the input of the equirectangular image. In addition, we remove the final fully connected layer thus retaining the image features extracted by the network.
\paragraph{Point Cloud Feature Extraction} The input point cloud is the set of points $\mathcal{P}=\left \{ p_1, p_2, \dots, p_N \mid  p_i \in \mathbb{R}^3 \right \}$. We employ the PointNet \cite{qi2017pointnet} network that can directly take in point clouds to identify features. To improve the generalization ability of the model and to satisfy the requirements of PointNet input, we first randomly select a fixed number of $\mathbf{N}_d=4096$ points from the point clouds submap before proceeding with the subsequent operations. Similarly, to retain the feature maps, we remove the last max pooling layer.

\subsubsection{Attention Enhancement}
After feature extraction, the data of the two modalities are separately mapped to the high-dimensional feature space. The feature map of the equirectangular image is $\mathbf{U}^I \in \mathbb{R}^{C_1\times H\times W}$, where $C_1$ represents the number of feature channels, $H$ and $W$ represent the image length and width after being downsampled by CNN. The feature map of point clouds is $\mathbf{U}^P \in \mathbb{R}^{C_2\times N\times 1}$, where $C_2$ represents the number of feature channels for point clouds and $N$ equals the input point number $\mathbf{N}_d$. $C_1$ is not necessarily equal to $C_2$ since the distribution and types of image and point cloud features are different. To capture those features of images and point clouds that are most salient to match two modal information in the same position. We use the squeeze-and-excitation method in \cite{hu2018squeeze} to perform \emph{feature re-calibration}. Specifically, assuming that $\mathbf{U}^I$ and $\mathbf{U}^P$ are denoted by $\mathbf{U} \in \mathbb{R}^{C\times D_1\times D_2}=\left \{ u_1, \dots, u_C\right \}$, in the squeeze operation, by aggregating the feature map $\mathbf{U}$ across their spatial dimensions (e.g. $H$ and $W$ dimension for images), a channel descriptor $\mathbf{Z}=\{z_1, \dots, z_C \}$ can be obtained. $z_i$ is calculated by:
\begin{equation}
   z_i = \frac{1}{D_1\times D_2}\sum_{i=1}^{D_2}\sum_{j=1}^{D_1}u_i(i,j) 
\end{equation}

To further learn a non-mutually-exclusive relationship \cite{hu2018squeeze} between features, in the second excitation operation, rescale factor $\mathbf{S}=\{s_1, \dots, s_C \}$ is calculated by:
\begin{equation}
  \mathbf{S} = \sigma ( \mathbf{W}_2 \delta ( \mathbf{W}_1\mathbf{Z}  )   ) 
\end{equation}
where $\mathbf{W}_1 \in \mathbb{R}^{\frac{C}{r}\times C}$ and $\mathbf{W}_2 \in \mathbb{R}^{C\times \frac{C}{r}}$ are the parameters of two fully connected layers with $r$ refers to the dimension reduction ratio, $\delta(\cdot) $ is the ReLU function and $\sigma(\cdot)$ refers to the sigmoid activation.

Eventually, the new feature map $\tilde{\mathbf{U}}=\{\tilde{u}_1, \dots, \tilde{u}_C \}$ after feature re-calibration can be obtained by:
\begin{equation}
  \tilde{u}_i = s_iu_i
\end{equation}

In this way, the network can adaptively learn the weights for different channels to selectively emphasize informative features and suppress less useful ones, which is valuable for further measuring the similarity between images and point clouds. Through attention enhancement, we get two more effective discriminative features, $\tilde{\mathbf{U}}^I \in \mathbb{R}^{C_1\times H\times W}$ for the image and $\tilde{\mathbf{U}}^P \in \mathbb{R}^{C_2\times N\times 1}$ for the point clouds.

\subsubsection{Global Description Aggregation}
Finally, we aggregate the local features into global features through the NetVLAD layer. Specifically, the aggregation layer calculates the sum of residuals of all local features denoted by ${\tilde{\mathbf{U}}} \in \mathbb{R}^{N\times C}$ with each clustering center $\mathbf{C} \in \mathbb{R}^{K\times C}=\{c_1, \dots, c_K \}$. The results are global features $\mathbf{G} \in \mathbb{R}^{K\times C}=\{g_1, \dots, g_K \}$, and $g_k$ is calculated by:
\begin{equation}
    g_k = \sum_{i=1}^{C}\frac{e^{\omega_{k}^{T}{\tilde{u}}_i + b_k}}{\textstyle \sum_{k'}e^{\omega_{k'}^{T}{\tilde{u}}_i + b_{k'}}}({\tilde{u}}_i - c_k)
\end{equation}
Where $\omega_{k}$ and $b_{k}$ represent the weights and biases of the contributions of different local features to the global features, both them are learnable parameters.

Further, since $\mathbf{G}^I \in \mathbb{R}^{K\times C_1}$ and $\mathbf{G}^P \in \mathbb{R}^{K\times C_2}$ are different in dimensions, we utilize the fully connected layer to compress each of them and then L2-normalize the result into a compact global descriptor $\mathbf{V} \in \mathbb{R}^{D}$. The reduced dimensions can also improve the efficiency of similarity computation between two modalities during actual retrieval.

In summary, on top of the NetVLAD layer that applies differentiable weights to local features to enable traditional VLAD learnable, our attention-enhanced approach further weights the channels of local features to make the model more capable of representation for cross-modal feature comparisons.

\subsubsection{Metric Learning}
Given an \emph{equirectangular image} and a point cloud \emph{sub-map}, the model outputs two global descriptors $\mathbf{V}^I$ and $\mathbf{V}^P$. To learn the functions $f(\cdot)$ and $g(\cdot)$ jointly, we obtain the image tuple $\{I_a, I_p, \{I_{negs}\}\}$ and the point cloud tuple $\{P_a, P_p, \{I_{negs}\}\}$ from the training set. $I_p$ refers to the positive sample image which depicts the same place as $I_a$, and $I_{negs}=\{I^1_{neg}, \dots, I^N_{neg}\}$ is a set of negative sample images taken in different places. The point cloud tuple is composed of corresponding point clouds.

The target of the loss function is to minimize the distance between $I_a$ and $P_p$ while maximizing the distance between $I_a$ and $\{P_{negs}\}$. So we construct the triplet $\{I_a, P_p, \{P_{negs}\}\}$, and the triplet loss is defined to keep the distance of $I_a$ from $P_p$ smaller than the distance of $I_a$ from $\{P_{negs}\}=\{P^1_{neg}, \dots, P^N_{neg}\}$:
\begin{equation}
    \mathcal{L}_{ItoP} = [d(f(I_a),g(P_p)) - d(f(I_a),g(P^i_{neg}))+m]_+
\end{equation}
where $d(\cdot)$ refers to the Euclidean distance function, $m$ refers to the margin of triplet loss function, and $[\cdot]_+$ means $Max([\cdot],0)$.

Likewise, given $P_a$, the distance between $P_a$ and $I_p$ is desired to be smaller than that with $\{I_{negs}\}$. Based on the triplet $\{P_a, I_p, \{I_{negs}\}\}$, the loss function is defined as:
\begin{equation}
    \mathcal{L}_{PtoI} = [d(f(P_a),g(I_p)) - d(f(P_a),g(I^i_{neg}))+m]_+
\end{equation}

Furthermore, to improve the ability of the two branching networks in extracting features of the respective modal data, based on the same-modal retrieval we construct $\{I_a, I_p, \{I_{negs}\}\}$ and $\{P_a, P_p, \{P_{negs}\}\}$ and the loss function is defined as:
\begin{equation}
\begin{aligned}
    &\mathcal{L}_{SM} = [d(f(I_a),g(I_p)) - d(f(I_a),g(I^i_{neg}))+m]_+\\
    &+ [d(f(P_a),g(P_p)) - d(f(P_a),g(P^i_{neg}))+m]_+\\
\end{aligned}
\end{equation}

The model can learn from the above loss to keep the distance of the anchor from the positive sample smaller than that from the negative samples. To directly compare the differences between the two modal representations of images and point clouds at the same location, we define the loss for anchor $I_a$ and $P_a$:
\begin{equation}
    \mathcal{L}_{anchor} = d(f(I_a),g(P_a))
\end{equation}

The resulting loss function $\mathcal{L}_{sum}$ is calculated by:
\begin{equation}
\begin{aligned}
    &\mathcal{L}_{CM} = \mathcal{L}_{ItoP}+\mathcal{L}_{PtoI}\\
    &\mathcal{L}_{sum} = \mu \mathcal{L}_{CM}+\lambda \mathcal{L}_{SM}+\nu \mathcal{L}_{anchor}
\end{aligned}
\end{equation}
where $\mu=1$, $\lambda=0.1$ and $\nu=1$ in practice. These parameters determine the weights of the three components.

\subsection{Implementation Detail}
Before being input into the network, the equirectangular image is resized to 512x1024. To speed up the convergence of the network and improve the generalization capability, it is then normalized to a distribution with zero mean and a standard deviation of 1. For the raw point cloud $\mathcal{P}$, we only utilize the (x,y,z) dimensions, thus $\mathcal{P}=\left \{ p_1, p_2, \dots, p_N \mid  p_i \in \mathbb{R}^3 \right \}$.

The whole implementation is performed with the PyTorch \cite{Paszke2019PyTorchAI} library. To prevent over-fitting while training the model, we perform augmentation on the images to simulate possible jitter, light changes, etc. in the actual process. The dimension of the output global descriptor is 256. For training, the batch size is set to 8. The model is trained for 50 epochs with an initial learning rate of 0.0001. The parameters of the 2D and 3D branching networks are simultaneously optimized by Stochastic Gradient Descent (SGD) and Adaptive Moment Estimation (Adam).


%% file: sections/experiments.tex
\section{Experiments}
\label{sec:experiments}
This section describes the dataset, the details, and the evaluation of our proposed method.
\subsection{Dataset}
Since we focus on the cross-modal localization between spherical images and point clouds, our method is trained and evaluated on the KITTI-360 Dataset \cite{liao2022kitti}. KITTI-360 is a large-scale driving dataset that contains dual fisheye images and push-broom laser scans captured by rich sensors, including a pair of fisheyes, a Velodyne HDL-64E, and a SICK laser scanning unit. KITTI-360 record about 80k frames of information (each frame corresponds to two perspective images, two fisheye images, and a laser scan) over a distance of 73.7km in several suburban areas. Besides, each frame has an accurate and geo-localized pose, which is essential to our task. Different from the RobotCar dataset \cite{Maddern20171Y1} used in \cite{uy2018pointnetvlad} \cite{cattaneo2020global}, which records the same trajectory multiple times a year, KITTI-360 is composed of different sequences, each corresponding to a continuous and rarely overlapped driving trajectory. We use eight of these sequences for training or evaluation. According to the number of frames in each sequence, we randomly select a continuous trajectory (1.5k frames) from \emph{s00 sequence} for evaluation and 500 to 1500 frames from each other sequence for further test. The rest data is used for training. 

\subsection{Data Preprocessing and Training Set Construction}
Since the original images are two \emph{185°} fisheye images captured with one fisheye camera to each side, we stitch the dual fisheye images to make an \emph{equirectangular image} as the representation of the 360 image. We build a global LiDAR map for each sequence by accumulating the scans of each trajectory according to the ground truth poses, and then cut out \emph{sub-maps} with richer scene information according to the local pose. For each sub-map, we remove the ground plane since it is non-informative and repetitive.

The ground truth pose is provided at a given frame if the moving distance from the last valid frame is larger than a threshold in \cite{liao2022kitti}. To avoid two contiguous frames being too similar when captured at a slow speed, we use the 360 image and point cloud \emph{sub-map} $\{\mathcal{I}, \mathbf{P}\}$ corresponding to each given pose. For the training set, we select an image every 3 meters as a query in the same sequence. The positive sample is randomly selected within a distance of fewer than \emph{20 meters}, and negative samples are randomly selected outside a distance of \emph{40 meters} from the query to increase data diversity.

\subsection{Evaluation Details}
As in \cite{cattaneo2020global}, the evaluation region $\mathcal{E}$ is a randomly selected continuous path from the \emph{s00 sequence}, and $\mathcal{E}$ never appears in the training set. To make each test more efficient, we select a query from $\mathcal{E}$ (1.5k frames) every \emph{10 meters} to avoid adjacent test queries being too close to each other. We consider the poses within a distance of \emph{20 meters} to be at the same location and evaluate the results based on the \emph{recall@k} method. If the top-k from the database has elements in the same position, then the localization is correct. The \emph{recall@k} counts the percentage of all queries that are correct. In addition, since it is more difficult to retrieve as the database increases, we also test \emph{recall@1\%} as a reference, in which \emph{k} equals 1\% of the database size.

\subsection{Result}
\label{exp:result}
\begin{figure}[t]
    \centering
    \includegraphics[width=\linewidth]{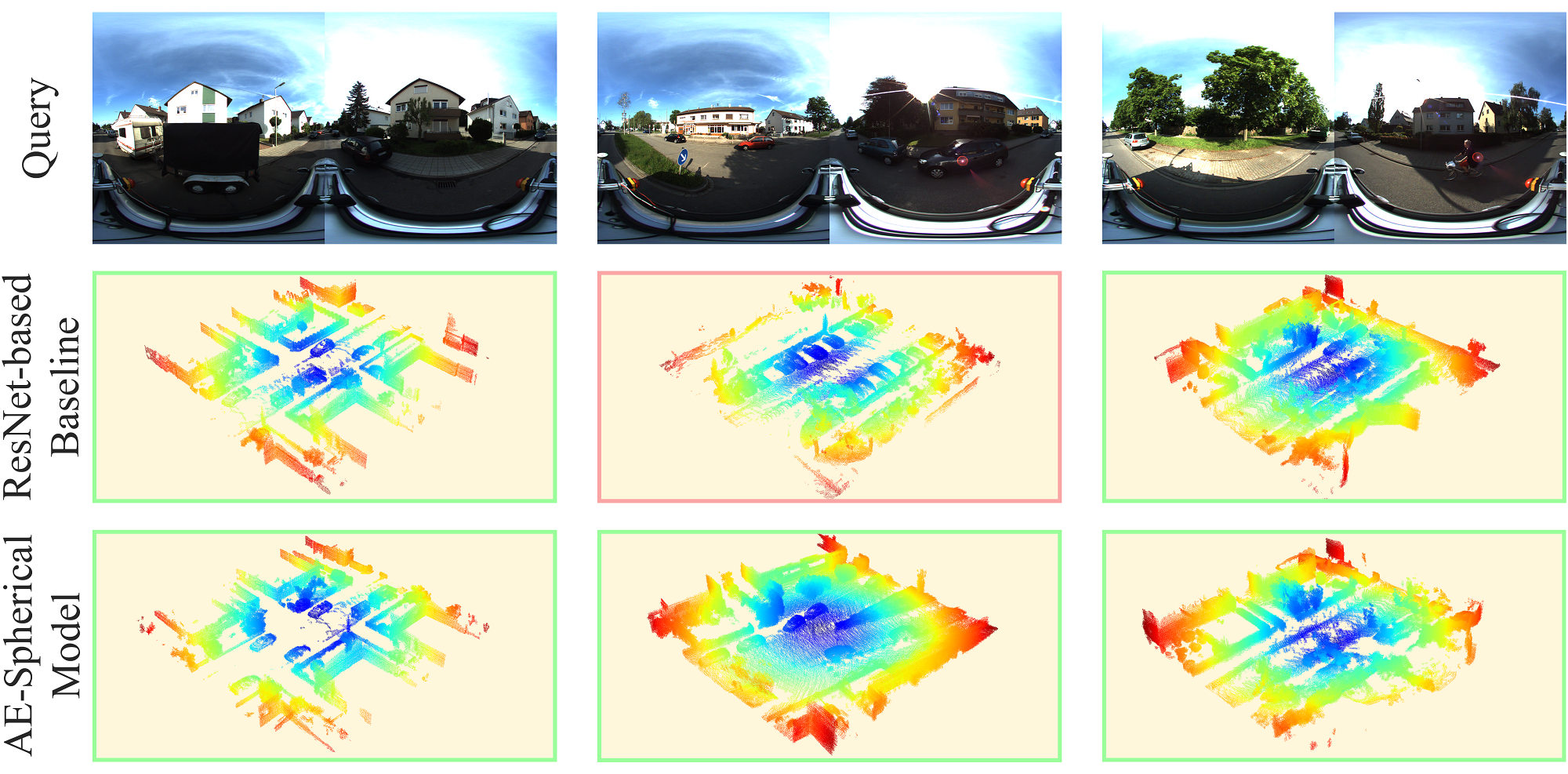}
    \caption{The results of cross-modal localization. The second and third rows show the recall@top1 point cloud sub-map retrieved through ResNet-based Baseline and AE-Spherical Model with the 360 image, where the green frame indicates a correct result and the red frame indicates an incorrect result.} \vspace{-3mm}
    \label{fig:result}
\end{figure}

Our ResNet-based baseline is composed of 2D network (ResNet-18 and NetVLAD) and 3D network (PointNet and NetVLAD), in which the features of equirectangular images are extracted by conventional 2D CNN. Our proposed refined model (AE-Spherical Model) employs ResNet-18-based \emph{spherical} CNN for feature extraction of equirectangular images and attention enhancement as described in detail in \ref{sec:approach}. The image feature extraction part of both models uses pre-trained weights for ResNet-18 offered by TorchVision (pre-trained on ImageNet\cite{DengImageNet} dataset).

\begin{table}[h!]
\caption{Comparison Between Models}
\centering
\begin{tabular}{ c c c c c }
\toprule
  model & recall@1 & recall@5 & recall@1\%\\
\midrule
 ResNet-based Baseline & 36.79 & 52.83 & 66.98  \\ 
 AE-Spherical Model & \textbf{46.23} & \textbf{66.04} & \textbf{75.47}\\
\bottomrule
\end{tabular}
\label{table:basevsref}
\end{table}

We compare AE-Spherical Model with the ResNet-based baseline and the results shown in TABLE \ref{table:basevsref} demonstrate that AE-Spherical Model outperforms the baseline method. Given 360 images, the accuracy of \emph{recall@1} improves by 9.44\%. It indicates that the refined model can better map 360 images and point clouds into a shared high-dimensional feature space. The refined model achieves improved \emph{recall@k} accuracy under different values of k.

To evaluate the cross-modal performance, we also perform the task of retrieving the 360 image with point clouds query for reference. We further evaluated the performance of the models on same-modal localization. Altogether, based on the modalities of the query and database data, four localization retrieval (2D-to-2D, 3D-to-3D, 2D-to-3D, 3D-to-2D),  tasks are performed with two models, AE-Spherical Model and ResNet-based baseline.

\begin{figure}[h!]
    \centering
    \includegraphics[width=\linewidth]{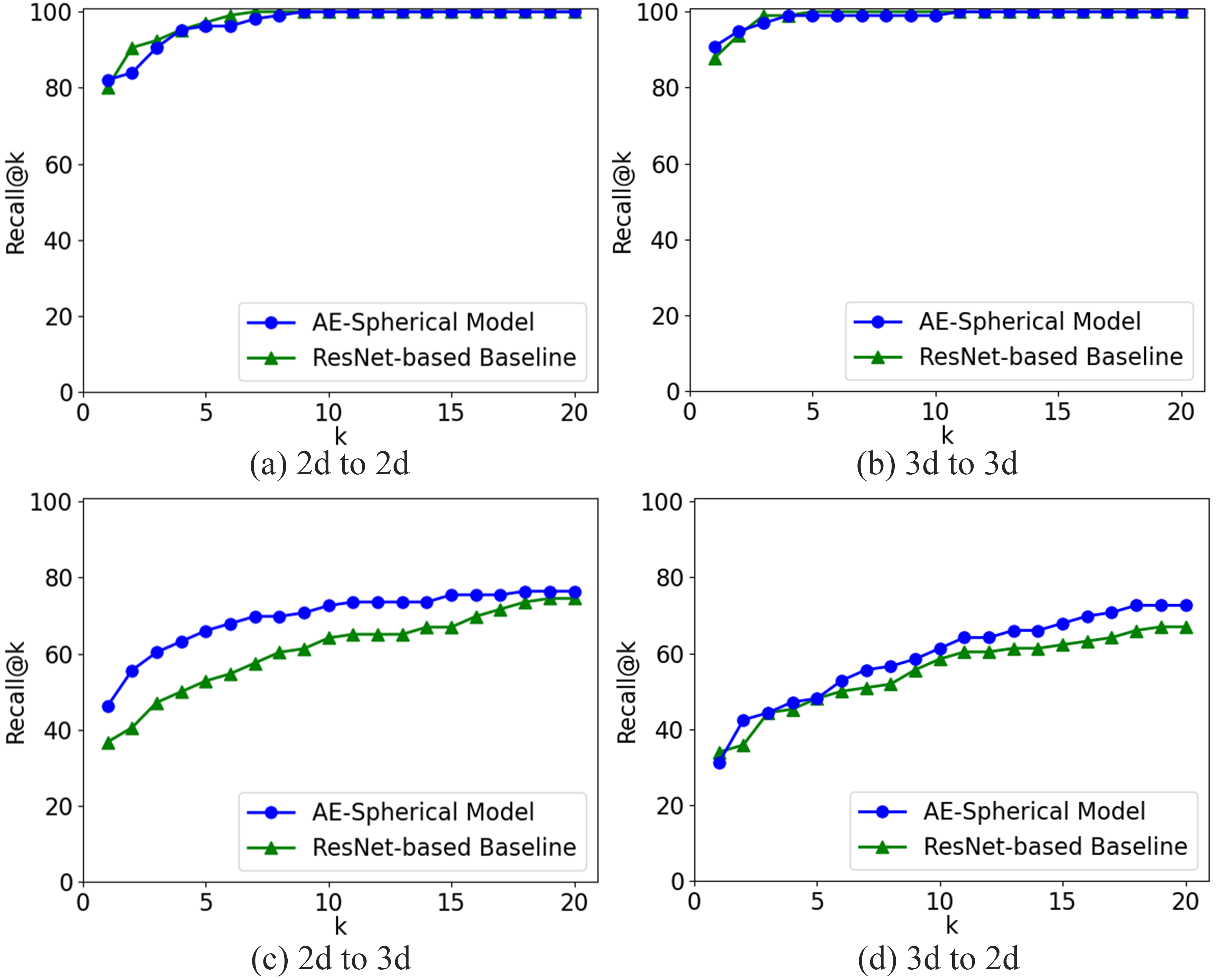}
    \caption{Recall@k measure of AE-Spherical Model and ResNet-based Baseline for four tasks including same-modal localization (a), (b) and cross-modal localization (c), (d).} \vspace{-3mm}
    \label{fig:recall_compare}
\end{figure}

The comparison results are shown in Fig.\ref{fig:recall_compare}. For the robotic system with localization task, it is better to retrieve the correct result in fewer \emph{top-k}. Therefore, we focus on the accuracy of \emph{recall@k} with k ranging from 1 to 20. From the cross-modal results, the overall performance of the AE-Spherical Model is better than the baseline, especially for 2D-to-3D when k is lower. Whereas, for the same-modal task, there is no significant difference in the performance of the two models, both being at a high level. The limited effect of attention on features can be explained by the small differences between features of same-modal data without heterogeneity. Besides, the retrieval between point clouds is little influenced by the spherical convolution for the 360 images.

\begin{table}[h!]
\centering
\caption{Comparison Between 2D CNN}
\begin{tabular}{ c c c | c c }
\toprule
& \multicolumn{2}{c}{recall@5} & \multicolumn{2}{c}{recall@1\%}  \\
\cmidrule{2-5}
  2D CNN & 2D to 3D & 3D to 2D & 2D to 3D & 3D to 2D\\
\midrule
 VGG-16 & 40.57 & 39.62 & 57.55 & 56.6 \\ 
 ResNet-18 & \textbf{42.45} & \textbf{40.57} & \textbf{59.43} & \textbf{62.26}\\
\bottomrule
\end{tabular}
\label{table:backbone}
\end{table}

\subsubsection{Comparison Between 2D CNN}
We first compare the effect of VGG-16 and ResNet-18 combined with attention enhancement to decide the backbone network for spherical CNN. The results in TABLE \ref{table:backbone} show that ResNet-18 performs better than VGG-16 when combined with attention enhancement in the task of cross-modal retrieval. As the first part of the model, later connected to the attention module and NetVLAD layer, ResNet-18 can better optimize the network to learn the features of the image during backpropagation with residual blocks.

\subsubsection{Comparison Between Losses}
We compare the effect of training our refined network with different loss functions. The results in TABLE \ref{table:loss} show the effectiveness of utilizing cross-modality $\mathcal{L}_{CM}$ as well as $\mathcal{L}_{anchor}$, and the model works best when trained with losses combined.

\begin{table}[h!]
\centering
\caption{Comparison Between Losses}
\setlength\tabcolsep{4.5pt}
\begin{tabular}{ c c c c c }
\toprule
  loss & recall@1 & recall@5 & recall@10 & recall@20\\
\midrule
0.1$\mathcal{L}_{SM}+\mathcal{L}_{anchor}$ & 16.04 & 33.96 & 41.51 & 57.55\\
 0.1$\mathcal{L}_{SM}+\mathcal{L}_{CM}$ & 28.30 & 42.45 & 57.55 & 67.92 \\ 
 $\mathcal{L}_{sum}$ & \textbf{37.74} & \textbf{53.77} & \textbf{59.43} & \textbf{70.75}\\
\bottomrule
\end{tabular}
\label{table:loss}
\end{table}

For the two-branch network, the loss function plays an important role in guiding the model to update toward learning effective features. When we remove $\mathcal{L}_{CM}$, the accuracy decreases a lot since the model cannot learn the representations for the cross-modal similarity measure by minimizing the distance of the query from positive samples and maximizing the distance from negative samples. The result demonstrates the effectiveness of $\mathcal{L}_{anchor}$ and the triplet loss used in $\mathcal{L}_{CM}$ for cross-modal learning.

\subsubsection{Ablation Study}
Further, We conduct an ablation study under the same condition as in TABLE \ref{table:basevsref} to investigate the effect of network components. In particular, we focus on the contribution of spherical CNN and attention enhancement. We build the model \emph{SCNN} based on the baseline model using the spherical convolution network instead of the conventional one. In addition, we apply attention enhancement on the basis of the baseline model to construct the model \emph{AE-ResNet}. 

\begin{table}[h!]
\centering
\caption{Ablation Study}
\setlength\tabcolsep{4.5pt}
\begin{tabular}{ c c c| c c | c c }
\toprule
\multicolumn{3}{c}{}& \multicolumn{2}{c}{recall@1} & \multicolumn{2}{c}{recall@1\%}  \\
\cmidrule{4-7}
 Base & \emph{SCNN} & \emph{Attention}& 2D to 3D & 3D to 2D & 2D to 3D & 3D to 2D\\
\midrule
 \Checkmark & \Checkmark & \XSolidBrush & 33.02 & 31.13 & 71.70 & 61.32 \\ 
 \Checkmark & \XSolidBrush & \Checkmark & 37.74 & 30.19 & 70.75 & \textbf{68.87}\\
 \Checkmark & \Checkmark & \Checkmark & \textbf{46.23} & \textbf{31.13} & \textbf{75.47} & 67.92\\
\bottomrule
\end{tabular}
\label{table:abla}
\end{table}

Since the pretrained parameters are based on ResNet-18 and have excellent feature extraction ability for images, \emph{SCNN} performs less well than baseline on recall@1 without \emph{Attention}. The results in TABLE \ref{table:abla} demonstrate the effectiveness of \emph{SCNN} for extracting distortion-invariant features and \emph{Attention} for focusing on salient features, which promote an overall improvement for cross-modal learning between 360 images and point clouds.


\subsubsection{Comparison Between Spherical Image and Perspective Image}
We train the model with perspective images under the same conditions and evaluate the effect of cross-modal localization with perspective images. In the KITTI360 \cite{liao2022kitti} dataset, the perspective image is captured at the same time as the spherical image, thus, we compare the effect of the perspective image and the spherical image on the cross-modal localization task. 

\begin{figure}[h!]
    \centering
    \includegraphics[width=\linewidth]{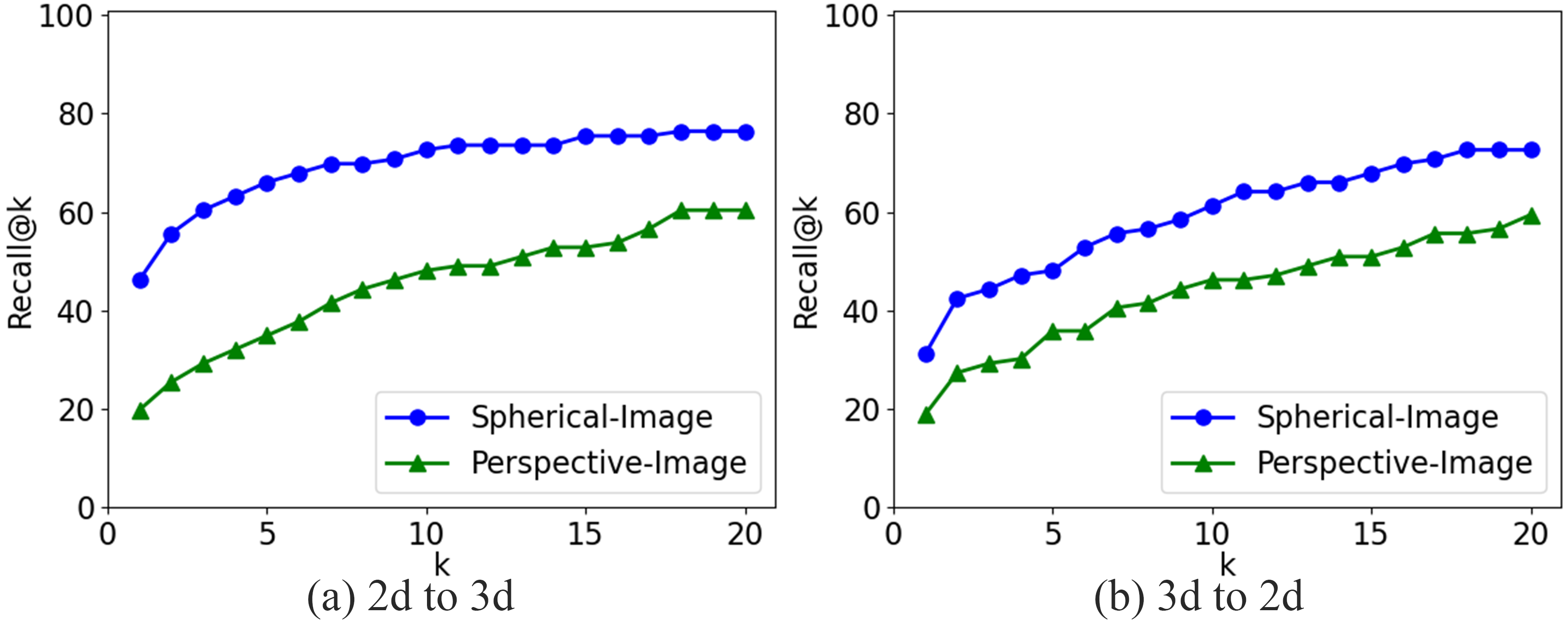}
    \caption{Recall@k measure Comparison between spherical image and perspective image for cross-modal localization tasks including (a) 2D to 3D, (b) 3D to 2D.} \vspace{-3mm}
    \label{fig:recall_perspective}
\end{figure}

The results in Fig.\ref{fig:recall_perspective} show that the spherical image performs significantly better in cross-modal localization. Since the spherical image can depict more comprehensive visual information and its omni-directional view corresponds to that of the submap, while perspective images can only acquire a limited range of image information, it is more effective to establish similarity between spherical images and point clouds.


\subsubsection{Further Test}
\label{exp:result:furtherTest}
KITTI360 \cite{liao2022kitti} records several suburban trajectories, where scenes can be divided into residential downtowns and highways. The images and point clouds of the downtown scenes contain information such as houses and greenfields, while the highway scenes contain relatively little information for distinguishing and recognizing, especially the point clouds with wide open areas. To further evaluate the model's performance in different scenarios, two test sets are constructed based on the characteristics of different trajectory sequences (data in the test set will not appear in training). Specifically, continuous downtown trajectories are selected from \emph{seq00}, \emph{seq02}, \emph{seq04}, \emph{seq06}, and \emph{seq09} and continuous highway trajectories from \emph{seq07} and \emph{seq10}, and then the average recall accuracy is measured respectively. 

\begin{figure}[h!]
    \centering
    \includegraphics[width=\linewidth]{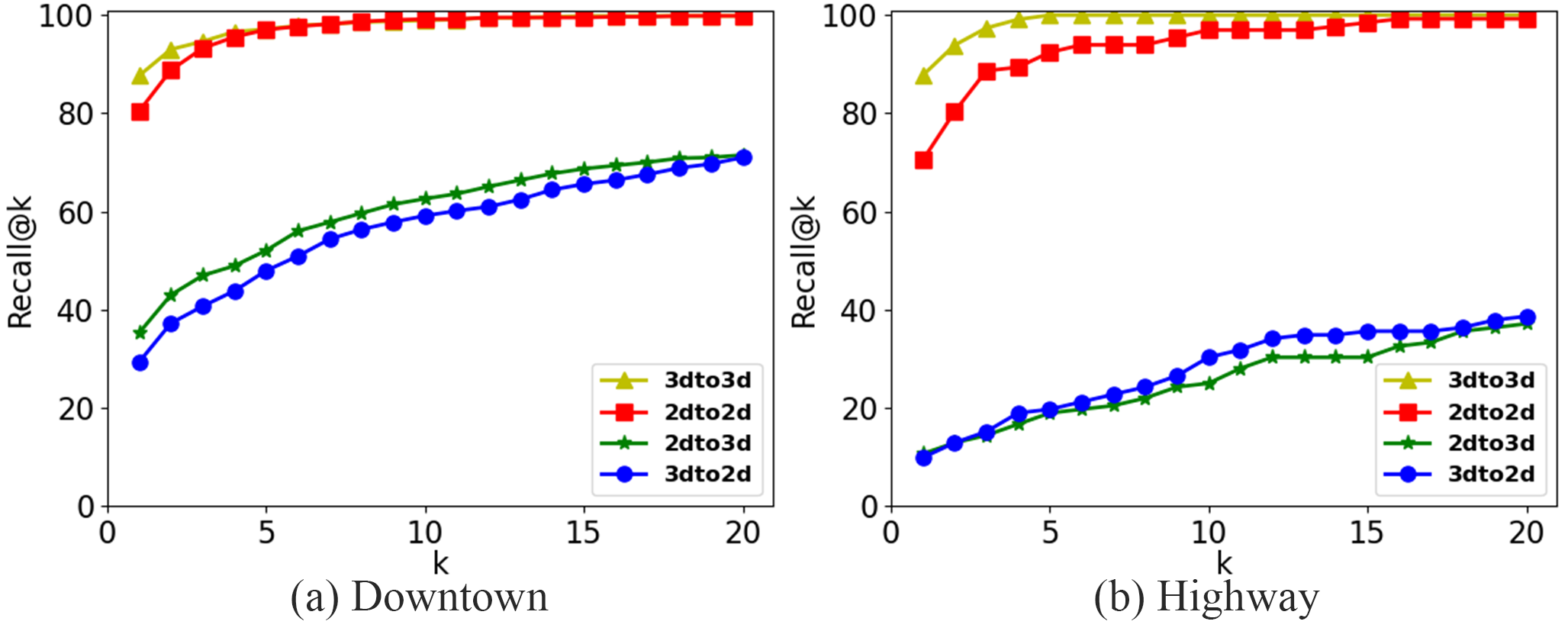}
    \caption{Average Recall@k of AE-Spherical Model on (a) Downtown and (b) Highway scenarios. The database for the downtown scenario test ranges from 1000 to 1500, and the database size for the road scenario test is 500 for both.} \vspace{-3mm}
    \label{fig:recall_expand}
\end{figure}
\vspace{-0mm}

The result of testing \emph{AE-Spherical} in two different scenarios is shown in Fig.\ref{fig:recall_expand}. For cross-modal localization, the model performs less well on highways than on downtown scenes. While the downtown has a variety of visual and geometric information, the highway tends to convey less semantic information, which is difficult to recognize even for human beings, and it makes the cross-modal retrieval accuracy at highways reduced.

%% file: sections/conclusion.tex
\section{Conclusions and Future Work}
\label{sec:conclusion}
In this paper, we propose a cross-modal trainable model for camera localization in LiDAR maps by performing retrieval between 360 images and point clouds. To bridge the cross-modal heterogeneity gap, we introduce an attention module that can guide the model to emphasize informative local features and suppress less important ones for both modalities by adaptively adjusting the weights. Besides, we utilize a spherical convolution network to mitigate the problem of distortion for 360 images. We train the model and conduct evaluations on the KITTI-360 dataset. With the significantly improved recall accuracy, we prove the advantage of spherical images over perspective images on cross-modal retrieval. Moreover, the comparison result between the refined model and the baseline model demonstrates that the optimization of the feature extraction and the attention on local features enable the model to better establish similarity in high-dimensional feature space. One notable limitation is that the model performs less well on highways. In the future, we intend to further enhance the coupling between the two branching networks to fully utilize limited semantic information.



%% file: main.bbl
\begin{thebibliography}{10}
\providecommand{\url}[1]{#1}
\csname url@rmstyle\endcsname
\providecommand{\newblock}{\relax}
\providecommand{\bibinfo}[2]{#2}
\providecommand\BIBentrySTDinterwordspacing{\spaceskip=0pt\relax}
\providecommand\BIBentryALTinterwordstretchfactor{4}
\providecommand\BIBentryALTinterwordspacing{\spaceskip=\fontdimen2\font plus
\BIBentryALTinterwordstretchfactor\fontdimen3\font minus
  \fontdimen4\font\relax}
\providecommand\BIBforeignlanguage[2]{{%
\expandafter\ifx\csname l@#1\endcsname\relax
\typeout{** WARNING: IEEEtran.bst: No hyphenation pattern has been}%
\typeout{** loaded for the language `#1'. Using the pattern for}%
\typeout{** the default language instead.}%
\else
\language=\csname l@#1\endcsname
\fi
#2}}

\bibitem{newman2005slam}
P.~Newman and K.~Ho, ``{SLAM-loop closing with visually salient features},'' in
  \emph{Proceedings of the IEEE International Conference on Robotics and
  Automation (ICRA)}, 2005, pp. 635--642.

\bibitem{yang2020mobile3drecon}
X.~Yang, L.~Zhou, H.~Jiang, Z.~Tang, Y.~Wang, H.~Bao, and G.~Zhang,
  ``{Mobile3DRecon}: real-time monocular 3d reconstruction on a mobile phone,''
  \emph{IEEE Transactions on Visualization and Computer Graphics}, vol.~26,
  no.~12, pp. 3446--3456, 2020.

\bibitem{tang2015neighborhood}
J.~Tang, Z.~Li, M.~Wang, and R.~Zhao, ``Neighborhood discriminant hashing for
  large-scale image retrieval,'' \emph{IEEE Transactions on Image Processing},
  vol.~24, no.~9, pp. 2827--2840, 2015.

\bibitem{arandjelovic2016netvlad}
R.~Arandjelovic, P.~Gronat, A.~Torii, T.~Pajdla, and J.~Sivic, ``{NetVLAD: CNN
  architecture for weakly supervised place recognition},'' in \emph{Proceedings
  of the IEEE conference on Computer Vision and Pattern Recognition (CVPR)},
  2016, pp. 5297--5307.

\bibitem{uy2018pointnetvlad}
M.~A. Uy and G.~H. Lee, ``{PointNetVLAD: Deep point cloud based retrieval for
  large-scale place recognition},'' in \emph{Proceedings of the IEEE conference
  on Computer Vision and Pattern Recognition (CVPR)}, 2018, pp. 4470--4479.

\bibitem{guo2019deep}
W.~Guo, J.~Wang, and S.~Wang, ``Deep multimodal representation learning: A
  survey,'' \emph{IEEE Access}, vol.~7, pp. 63\,373--63\,394, 2019.

\bibitem{feng20192d3d}
M.~Feng, S.~Hu, M.~H. Ang, and G.~H. Lee, ``{2D3D-MatchNet: Learning to match
  keypoints across 2d image and 3d point cloud},'' in \emph{Proceedings of the
  IEEE International Conference on Robotics and Automation (ICRA)}, 2019, pp.
  4790--4796.

\bibitem{cattaneo2020global}
D.~Cattaneo, M.~Vaghi, S.~Fontana, A.~L. Ballardini, and D.~G. Sorrenti,
  ``Global visual localization in lidar-maps through shared 2d-3d embedding
  space,'' in \emph{Proceedings of the IEEE International Conference on
  Robotics and Automation (ICRA)}, 2020, pp. 4365--4371.

\bibitem{courbon2007generic}
J.~Courbon, Y.~Mezouar, L.~Eckt, and P.~Martinet, ``A generic fisheye camera
  model for robotic applications,'' in \emph{IEEE/RSJ International Conference
  on Intelligent Robots and Systems (IROS)}, 2007, pp. 1683--1688.

\bibitem{Coors2018ECCV}
B.~Coors, A.~P. Condurache, and A.~Geiger, ``{SphereNet}: Learning spherical
  representations for detection and classification in omnidirectional images,''
  in \emph{Proceedings of the European Conference on Computer Vision (ECCV)},
  2018, pp. 518--533.

\bibitem{liao2022kitti}
Y.~Liao, J.~Xie, and A.~Geiger, ``{KITTI-360: A novel dataset and benchmarks
  for urban scene understanding in 2d and 3d},'' \emph{IEEE Transactions on
  Pattern Analysis and Machine Intelligence}, 2022.

\bibitem{haralick1991analysis}
R.~M. Haralick, C.-n. Lee, K.~Ottenburg, and M.~N{\"o}lle, ``Analysis and
  solutions of the three point perspective pose estimation problem.'' in
  \emph{Proceedings of the IEEE conference on Computer Vision and Pattern
  Recognition (CVPR)}, vol.~91, 1991, pp. 592--598.

\bibitem{lee2014unsupervised}
G.~H. Lee and M.~Pollefeys, ``Unsupervised learning of threshold for geometric
  verification in visual-based loop-closure,'' in \emph{Proceedings of the IEEE
  International Conference on Robotics and Automation (ICRA)}, 2014, pp.
  1510--1516.

\bibitem{lowe2004distinctive}
D.~G. Lowe, ``Distinctive image features from scale-invariant keypoints,''
  \emph{International Journal of Computer Vision}, vol.~60, no.~2, pp. 91--110,
  2004.

\bibitem{Bay2006SURF}
H.~Bay, T.~Tuytelaars, and L.~V. Gool, ``{SURF}: Speeded up robust features,''
  in \emph{Proceedings of the European Conference on Computer Vision (ECCV)},
  2006, pp. 404--417.

\bibitem{Rublee2011ORB}
E.~Rublee, V.~Rabaud, K.~Konolige, and G.~Bradski, ``{ORB}: An efficient
  alternative to sift or surf,'' in \emph{International Conference on Computer
  Vision (ICCV)}, 2011, pp. 2564--2571.

\bibitem{simonyan2014very}
K.~Simonyan and A.~Zisserman, ``Very deep convolutional networks for
  large-scale image recognition,'' in \emph{International Conference on
  Learning Representations (ICLR)}, 2015.

\bibitem{he2016deep}
K.~He, X.~Zhang, S.~Ren, and J.~Sun, ``Deep residual learning for image
  recognition,'' in \emph{Proceedings of the IEEE conference on Computer Vision
  and Pattern Recognition (CVPR)}, 2016, pp. 770--778.

\bibitem{s.2018spherical}
T.~S. Cohen, M.~Geiger, J.~K{\"o}hler, and M.~Welling, ``Spherical {CNN}s,'' in
  \emph{International Conference on Learning Representations (ICLR)}, 2018.

\bibitem{nister2006scalable}
D.~Nister and H.~Stewenius, ``Scalable recognition with a vocabulary tree,'' in
  \emph{IEEE Computer Society Conference on Computer Vision and Pattern
  Recognition (CVPR)}, vol.~2, 2006, pp. 2161--2168.

\bibitem{jegou2010aggregating}
H.~J{\'e}gou, M.~Douze, C.~Schmid, and P.~P{\'e}rez, ``Aggregating local
  descriptors into a compact image representation,'' in \emph{IEEE Computer
  Society Conference on Computer Vision and Pattern Recognition (CVPR)}, 2010,
  pp. 3304--3311.

\bibitem{qi2017pointnet}
C.~R. Qi, H.~Su, K.~Mo, and L.~J. Guibas, ``{PointNet}: Deep learning on point
  sets for 3d classification and segmentation,'' in \emph{Proceedings of the
  IEEE conference on Computer Vision and Pattern Recognition (CVPR)}, 2017, pp.
  652--660.

\bibitem{zhong2009intrinsic}
Y.~Zhong, ``Intrinsic shape signatures: A shape descriptor for 3d object
  recognition,'' in \emph{IEEE 12th International Conference on Computer Vision
  workshops}, 2009, pp. 689--696.

\bibitem{wang2019dynamic}
Y.~Wang, Y.~Sun, Z.~Liu, S.~E. Sarma, M.~M. Bronstein, and J.~M. Solomon,
  ``Dynamic graph cnn for learning on point clouds,'' \emph{ACM Transactions on
  Graphics}, vol.~38, no.~5, pp. 1--12, 2019.

\bibitem{yan2018second}
Y.~Yan, Y.~Mao, and B.~Li, ``{SECOND}: Sparsely embedded convolutional
  detection,'' \emph{Sensors}, vol.~18, no.~10, p. 3337, 2018.

\bibitem{yin2021i3dloc}
P.~Yin, L.~Xu, J.~Zhang, H.~Choset, and S.~Scherer, ``i3dloc: Image-to-range
  cross-domain localization robust to inconsistent environmental conditions,''
  in \emph{Robotics: Science and Systems (RSS)}, 2021.

\bibitem{lai2022adafusion}
H.~Lai, P.~Yin, and S.~Scherer, ``Adafusion: Visual-lidar fusion with adaptive
  weights for place recognition,'' \emph{IEEE Robotics and Automation Letters},
  vol.~7, no.~4, pp. 12\,038--12\,045, 2022.

\bibitem{hu2018squeeze}
J.~Hu, L.~Shen, and G.~Sun, ``Squeeze-and-excitation networks,'' in
  \emph{Proceedings of the IEEE conference on Computer Vision and Pattern
  Recognition (CVPR)}, 2018, pp. 7132--7141.

\bibitem{Paszke2019PyTorchAI}
A.~Paszke, S.~Gross, F.~Massa, A.~Lerer, J.~Bradbury, G.~Chanan, T.~Killeen,
  Z.~Lin, N.~Gimelshein, L.~Antiga, \emph{et~al.}, ``{PyTorch}: An imperative
  style, high-performance deep learning library,'' \emph{Advances in Neural
  Information Processing Systems}, vol.~32, 2019.

\bibitem{Maddern20171Y1}
W.~Maddern, G.~Pascoe, C.~Linegar, and P.~Newman, ``1 year, 1000 km: The oxford
  robotcar dataset,'' \emph{The International Journal of Robotics Research},
  vol.~36, no.~1, pp. 3--15, 2017.

\bibitem{DengImageNet}
J.~Deng, W.~Dong, R.~Socher, L.-J. Li, K.~Li, and L.~Fei-Fei, ``{ImageNet}: A
  large-scale hierarchical image database,'' in \emph{Proceedings of the IEEE
  conference on Computer Vision and Pattern Recognition (CVPR)}, 2009, pp.
  248--255.

\end{thebibliography}
